\documentclass[lettersize,journal]{IEEEtran}
\usepackage{amsmath,amsfonts}
\usepackage{algcompatible}
\usepackage{algorithm}
\usepackage{array}

\usepackage{stackengine}

\usepackage{amssymb}
\usepackage{textcomp}
\usepackage{stfloats}
\usepackage{url}
\usepackage{verbatim}
\usepackage{graphicx}
\usepackage{xcolor}
\usepackage{cite}
\usepackage{amsmath}
\usepackage{tabularx}
\usepackage[english]{babel}
\usepackage{amsthm}
\usepackage{multirow}
\usepackage{booktabs}
\usepackage{mathtools}
\usepackage{subcaption}
\usepackage{caption}
\usepackage{hyperref}

\def\A{U}
\def\B{V}

\def\AA{\Lambda}

\def\BibTeX{{\rm B\kern-.05em{\sc i\kern-.025em b}\kern-.08em
    T\kern-.1667em\lower.7ex\hbox{E}\kern-.125emX}}

\begin{document}

\title{
Predicting Power-System Dynamic Trajectories with Foundation Models
}

\author{
Haoran~Li,~\IEEEmembership{Member,~IEEE,}
Lihao~Mai,~\IEEEmembership{Student Member,~IEEE,}
Chenhan Xiao,~\IEEEmembership{Student Member,~IEEE,}
Erik Blasch,~\IEEEmembership{Fellow,~IEEE,}
and~Yang~Weng,~\IEEEmembership{Senior Member,~IEEE}
\thanks{
Haoran Li, Lihao Mai, Chenhan Xiao, and Yang Weng are with the School of Electrical, Computer and Energy Engineering at Arizona State University, Tempe, AZ 85281 USA, e-mail: \{lhaoran, lmai7, cxiao20, yang.weng\}@asu.edu.

Erik Blasch is with the MOVEJ Analytics, Dayton, OH 45324 USA, e-mail: erik.blasch@gmail.com.}
\vspace{-12mm}
}

\maketitle

\begin{abstract}
As power systems transition toward renewable-rich and inverter-dominated operations, accurate time-domain dynamic analysis becomes increasingly critical. Such analysis supports key operational tasks, including transient stability assessment, dynamic security analysis, contingency screening, and post-fault trajectory evaluation. In practice, these tasks may operate under several challenges, including unknown and time-varying system parameters, privacy constraints on data sharing, and the need for fast online inference. Conventional high-fidelity prediction based on nonlinear differential–algebraic equations (DAEs) requires detailed domain knowledge and can be computationally expensive. Meanwhile, existing learning-based approaches are typically trained for individual systems and therefore lack generalization across operating conditions and physical parameters. 
Hence, this paper proposes LArge Scale Small ODE (LASS)-ODE-Power, a learning framework for general-purpose time-domain prediction. The proposed approach leverages large-scale pretraining on more than 40 GB of DAE or ordinary differential-equation (ODE) trajectories to learn transferable representations. The resulting model supports trajectory prediction from short measurement prefixes across diverse dynamic regimes, including electromechanical and inverter-driven systems. Hence, the model can be directly used without data sharing in a zero-shot setting. In addition, the proposed architecture incorporates parallel and linearized computation to achieve fast inference. Moreover, to enhance task-specific performance in power systems, a specialized fine-tuning strategy is developed based on approximately 1 GB of heterogeneous power-system dynamic data. Extensive experiments over diverse power-system simulation scenarios demonstrate that LASS-ODE-Power consistently outperforms existing learning-based models in trajectory prediction accuracy with efficient inference.
\end{abstract}


\vspace{-6mm}
\section{Introduction}
\vspace{-2mm}
The electric power grid is undergoing a structural transition driven by high penetrations of inverter-based resources (IBRs) \cite{ref:Hamad2026B}, expanding AC/DC interconnections \cite{Pedra2021ThreePort}, and increasingly variable operating conditions \cite{li2025external}. These changes reduce operational margins and increase the need for accurate time-domain dynamic analysis. Such analysis supports several critical tasks in power system operation, including transient stability assessment \cite{Lara2024TDSim,Sobbouhi2021TSReview}, dynamic security assessment (DSA) \cite{Ren2022OnlineDSA}, and dynamic state estimation \cite{ref:Zhihao2024H,ref:Lihao2025G,ref:Haoran2025L}. It is also important for contingency screening \cite{Yogarathnam2024ContingencyIndex} and post-fault trajectory analysis \cite{ref:Haoran2019H,ref:Haoran2019U,ref:Haoran2022T}. In modern grid operations, such simulations must be executed repeatedly across many operating points, topology configurations, and disturbance scenarios under strict time constraints \cite{Ren2022OnlineDSA,Lara2024TDSim,ref:Haoran2024D}. As a result, scalable and efficient time-domain simulation and prediction have become increasingly important for both planning and real-time operational studies.

In practice, this goal introduces several challenges. First, dynamic-model parameters are often uncertain or time-varying due to changing operating conditions, controller updates, etc. In addition, fault events, protection actions, and switching operations induce discrete mode transitions that produce piecewise hybrid trajectories that are difficult to capture using a single fixed-parameter model \cite{XiaoSmartGridComm,li2025neural}. Second, measurement data-quality issues are pervasive, including noise, missing channels, asynchronous sampling, and partial observability, which can degrade numerical stability and the generalization performance of learned surrogates \cite{10086650,li2025latent,10891664}. Third, privacy and proprietary constraints often prevent operators and vendors from sharing measurements and device parameters in centralized training environments \cite{cui2025privacy}. Finally, practical studies require evaluating large ensembles of disturbance scenarios and operating conditions within tight latency budgets, making high-throughput and fast inference essential \cite{cheng2022pitpis,cao2023ftrt}.

Mathematically, time-domain dynamic prediction corresponds to solving nonlinear initial-value problems described by differential-algebraic equations (DAEs) that couple device-level differential dynamics with algebraic network constraints. In renewable-rich and inverter-dominated grids, the resulting dynamics exhibit pronounced multi-timescale behavior and hybrid switching events. Practical studies may therefore rely on electromagnetic-transient (EMT) simulation or EMT-transient-stability (TS) hybrid/co-simulation with explicit event handling \cite{subedi2021emtaccel,lin2021adaptive}. Increasingly, such simulations are aligned with field measurements to capture real-time system dynamics and calibrate models using heterogeneous monitoring streams with different sampling rates and resolutions \cite{mai2024data,10636963}.

Classic high-fidelity numerical simulation and prediction remain the gold standard for analyzing power-system dynamics when accurate parameters and models are available. For example, EMT simulation and EMT-TS co-simulation provide highly detailed representations of dynamic behavior \cite{konara2023cosim}. However, these approaches are computationally expensive because they repeatedly solve nonlinear network equations and perform fine-step numerical integration. Consequently, evaluating large ensembles of disturbance scenarios can become throughput-limited even when accelerated using GPU or high-performance computing techniques \cite{xiong2024paraemt}. Moreover, uncertain parameters, repeated initialization procedures, and time-varying operating conditions can further degrade simulation fidelity.
To reduce computational costs and improve modeling under uncertain parameters, learning-based approaches for dynamic prediction have recently attracted attention. Symbolic regression methods attempt to recover governing equations directly from trajectory data, for example, using genetic programming \cite{schmidt2009distilling} or sparse identification techniques such as SINDy \cite{brunton2016discovering}. While these approaches can produce interpretable models when the search space is well designed and data quality is high, they are often sensitive to noise, missing measurements, and derivative estimation errors, and they scale poorly to large and high-dimensional systems.

Another major family of methods is physics-informed learning, such as physics-informed neural networks (PINNs), which enforce DAE residual constraints during training to promote physically consistent prediction and parameter estimation \cite{raissi2019physics,karniadakis2021physics}. Several recent studies have explored such techniques in power-system applications \cite{ref:Haoran2025L,ref:Lihao2025G,10415845}. However, enforcing physical consistency typically requires partial knowledge of the underlying DAE structure, which can limit portability across heterogeneous devices and modeling conventions. Closely related continuous-time approaches, such as neural differential equation models \cite{chen2018neural,rackauckas2020universal}, parameterize system dynamics using neural networks and learn them through differentiable ODE solvers. Although these models can approximate complex nonlinear dynamics without requiring explicit parameters, training and inference remain computationally expensive due to repeated numerical integration. In addition, such models are often trained for individual systems, making cross-system generalization difficult.

These limitations motivate the development of reusable learning models that can generalize across systems, remain robust under realistic measurement imperfections, preserve privacy constraints, and enable high-throughput time-domain prediction without costly nonlinear integration. Achieving these goals simultaneously is challenging, particularly when models are trained on system-specific datasets with limited coverage of dynamic behaviors.
A promising direction is to learn transferable representations of dynamic trajectories from large historical datasets. Inspired by recent advances in large-scale pretraining across machine learning domains, our prior work proposed \underline{LA}rge \underline{S}cale \underline{S}mall ODE (LASS-ODE) \cite{li2026lass}, a reusable model for learning dynamic systems from diverse ODE/DAE trajectories. LASS-ODE supports two key capabilities: online parameter adaptation and trajectory prediction conditioned on short observed measurement segments. Through large-scale pretraining on more than 40\,GB of DAE trajectories, the model learns general dynamic patterns and improves robustness under measurement noise and missing data. In addition, the architecture incorporates linearized and parallel computation to enable efficient training and inference.

In this work, we instantiate this paradigm for power systems and develop \textit{LASS-ODE-Power}, a foundation model for measurement-based adaptive time-domain prediction in power grids. The pretrained LASS-ODE model is further fine-tuned on approximately 1 GB of multi-timescale power-system dynamic data to adapt it to grid-specific dynamics. A widely used adaptation strategy is low-rank adaptation (LoRA) \cite{hu2022lora}, which fine-tunes pretrained models using domain-specific data. However, directly applying LoRA to heterogeneous power-system dynamics often produces limited or even negative improvement, because the underlying dynamic processes vary substantially across operating regimes and simulation types. To overcome this limitation, we propose a clustering-based mixture-of-LoRA approach that partitions the data into more homogeneous clusters and assigns specialized low-rank adaptations to each cluster. This LoRA-clustering design preserves regime-specific information more effectively and avoids the performance degradation caused by naively aggregating highly heterogeneous dynamics. We evaluate the resulting model on a range of prediction tasks, including frequency/voltage stability and EMT simulations, etc. The results demonstrate accurate trajectory prediction, robustness to measurement imperfections and resolution changes, strong zero-shot transfer across scenarios, and fast inference suitable for large-scale what-if analysis.

The main contributions of this paper are threefold: We (1) present the LASS-ODE-Power fine-tuning and deployment pipeline for measurement-conditioned power-system dynamic prediction; (2) demonstrate consistent accuracy and robustness across diverse scenarios compared with strong learning-based baselines; and (3) evaluate runtime performance and show that the proposed approach enables high-throughput inference.

The rest of the paper is organized as follows. Section~\ref{sec:problem} formulates the prediction problem. Section~\ref{sec:lassode} presents the LASS-ODE formulation and architecture. Section~\ref{sec:mix_lora} demonstrates the mixture of LoRA technology to fine-tune LASS-ODE and obtain more specialized LASS-ODE-Power for power systems. Section~\ref{sec:power_app} describes the power-system tasks and trajectory dataset construction. Section~\ref{sec:theorem} gives computational analysisf to support the fast inference of LASS-ODE-Power. Section~\ref{sec:results} reports numerical results and ablation studies. Section~\ref{sec:conclusion} concludes the paper and discusses future research directions.

\vspace{-6mm}

\section{Problem Formulation}
\label{sec:problem}

\vspace{-2mm}

Power-system dynamics are commonly described by nonlinear differential-algebraic equations (DAEs) of the form
\begin{equation}
\dot{\boldsymbol{x}}_d = \boldsymbol{f}(\boldsymbol{x}_d, \boldsymbol{x}_a), 
\qquad
\boldsymbol{0} = \boldsymbol{g}(\boldsymbol{x}_d, \boldsymbol{x}_a),
\end{equation}
where $\boldsymbol{x}_d \in \mathbb{R}^{n_d}$ collects the differential states and $\boldsymbol{x}_a \in \mathbb{R}^{n_a}$ collects the algebraic states. Here, $\boldsymbol{f}(\cdot)$ characterizes the dynamic evolution of system components, whereas $\boldsymbol{g}(\cdot)$ enforces the algebraic constraints induced by the electrical network and device interconnections. Typical entries of $\boldsymbol{x}_d$ include generator rotor angles and speeds, transient internal voltages, exciter and governor states, and inverter control states, while $\boldsymbol{x}_a$ typically includes bus voltage magnitudes, voltage phase angles, injected currents, and other network variables determined through instantaneous algebraic constraints.

\begin{itemize}
    \item \textbf{Problem:} Develop a universal predictor, powered by foundation artificial intelligence (AI) models, for time-domain simulation of modern power systems governed by nonlinear DAEs. The predictor should generalize across heterogeneous network configurations, operating conditions, disturbance scenarios, and measurement patterns.
    \item \textbf{Given:} For simplicity, let the full system state be denoted by $\boldsymbol{x}(t)=\big[\boldsymbol{x}_d(t),\boldsymbol{x}_a(t)\big]\in\mathbb{R}^{d_x},$ where \(\boldsymbol{x}_d(t)\) and \(\boldsymbol{x}_a(t)\) denote the differential and algebraic states, respectively. We are given a collection of observed trajectory segments $\{\boldsymbol{x}_i(t_j^i)\}_{i,j},$ where \(i\) indexes different system instances and $\mathcal{T}_{\mathrm{obs}}^i:=\{t_j^i\}_{j=1}^{N_i}$ is the set of observation timestamps for the $i^{th}$  instance. The state dimension, observation horizon, and temporal resolution may vary across instances.
    \item \textbf{Find:} Learn a general-purpose foundation model that, from a short observed prefix of each trajectory, infers the underlying dynamic structure, interpolates the observed segments, and predicts the future evolution. Basically, the model should reconstruct the whole trajectory $\hat{\boldsymbol{x}}_i(t), t\in[0,t_{\max}^i],t_{\max}^i>t_{N_i}^i$, where $t_{\max}^i$ is a manually-defined maximal time of the trajectory.
\end{itemize}

In the following, we omit the system index \(i\) for notational simplicity. Nevertheless, the proposed foundation AI model is pretrained and fine-tuned on trajectory data drawn from a diverse collection of systems.

\begin{figure*}[t]
\centering
\includegraphics[width=\linewidth]{bigpic.pdf}
\caption{\textbf{Top}: Overview of the proposed framework, illustrating the pretraining and fine-tuning stages. \textbf{Bottom}: Architecture of LASS-ODE. Key components include GRU (Gated Recurrent Unit), RBF (Radial Basis Function), CSH (Common Structure Hub), Norm (Layer Normalization), MHA (Multi-Head Attention), and MoE (Mixture of Experts).}
\label{fig:bigpic}
\vspace{-8mm}
\end{figure*}

\vspace{-4mm}
\section{Preliminaries of LASS-ODE}
\vspace{-2mm}

\label{sec:lassode}

LASS-ODE is a general-purpose foundation architecture for ODE/DAE systems. It aims to learn transferable dynamic representations from large collections of heterogeneous trajectories while remaining consistent with continuous-time evolution. As shown in the bottom of Fig. \ref{fig:bigpic}, LASS-ODE contains three main components: a unified interface for heterogeneous trajectories, an attention-based backbone for reusable dynamic structure, and an ODE-constrained decoder for continuous-time reconstruction. More details can be found in \cite{li2026lass}.

\noindent\textbf{Unified handling of heterogeneous trajectories.}
The top left of Fig. \ref{fig:bigpic} shows datasets from different dynamical systems (around 40 GB). Pre-training on these data enables LASS-ODE to aggregate ODE/DAE knowledge from diverse systems and improve time-domain dynamic simulation. A core challenge is that trajectories may differ in state dimension, operating condition, topology, and temporal resolution.

To improve flexibility, LASS-ODE uses \emph{channel-independent} preprocessing: each state channel is first encoded independently, while cross-channel dependencies are modeled later in the backbone. State values are normalized to a common range, and time is rescaled to a normalized interval to reduce disparities across amplitudes and timescales. For systems with different timescales, the maximum time used for normalization also differs. The prediction task is prefix-conditioned extrapolation: given a short observed prefix over $[0,\tilde{t}_{N}]$ with $\tilde{t}_{N}<1$, the model reconstructs the full trajectory over $[0,1]$.

\noindent\textbf{Trajectory tokenization with temporal coherence.}
Here, \emph{tokenization} converts a continuous-time trajectory into temporally ordered local segments, while \emph{embedding} maps each segment to a learnable feature vector for later attention-based processing. Tokenization is important because it defines the elementary units for pattern discovery, and embeddings determine how these units are represented for comparison and composition.

Unlike standard tokenization in many foundation models \cite{kaplan2020scaling,liu2025sundial}, ODE/DAE trajectories require temporal coherence, since they are governed by an underlying function over time. For channel \(j\), a time-aware gated recurrent unit (GRU) first summarizes the observed prefix:
\begin{equation}
\label{eq:gru_encoder}
\boldsymbol{h}^{(j)}=\mathrm{GRU}\!\left(\{[\boldsymbol{x}^{(j)}(t_i);\Delta t_i]\}_{i=0}^{N}\right),
\end{equation}
where \(\Delta t_i\) is the time increment. We include \(\Delta t_i\) explicitly because it may vary across systems. The hidden feature \(\boldsymbol{h}^{(j)}\) captures global temporal information for channel \(j\).

LASS-ODE then partitions the normalized interval into \(K_{\text{token}}\) segments and localizes each token using a radial basis function (RBF)-based time embedding with a multi-layer perceptron (MLP):
\begin{equation}
\begin{aligned}
(\boldsymbol{\gamma}_k;\boldsymbol{\beta}_k)
&=
\mathrm{MLP}\!\big(\mathrm{LN}(\phi_{\mathrm{rbf}}(t_k^{\mathrm{start}}))\big), \\
\boldsymbol{e}_{jk}
&=
\boldsymbol{\gamma}_k \odot \mathrm{MLP}(\boldsymbol{h}^{(j)})+\boldsymbol{\beta}_k ,
\end{aligned}
\end{equation}
where \(\phi_{\mathrm{rbf}}(t_k^{\mathrm{start}})\) is the RBF embedding of the token start time, $\mathrm{LN}(\cdot)$ denotes layer normalization, and $\gamma_k,\beta_k\in\mathbb{R}^{d_{\text{model}}}$ are token-dependent scale and shift vectors, respectively.
Here, \(\boldsymbol{e}_{jk}\in\mathbb{R}^{d_{\text{model}}}\) is the embedding for channel \(j\) and token \(k\). A learnable channel encoding is added to distinguish state variables. Thus, each token embedding contains channel identity, local temporal position, and trajectory-level temporal context. Stacking embeddings across \(d_x\) channels and \(K_{\text{token}}\) tokens yields \(\tilde{E}\in\mathbb{R}^{K_{\text{token}}d_x \times d_{\text{model}}}\) (bottom middle of Fig. \ref{fig:bigpic}).

\noindent\textbf{Inter- and Intra-Attention to Extract High-Level Features.}
Attention is central because it adaptively determines which tokens should interact, allowing the model to integrate information across distant positions and channels and extract reusable higher-level structure. After tokenization, all channel-time tokens are stacked and passed to a Transformer-style backbone.

Previous methods mainly use intra-system attention, where interactions are restricted to one system instance. While effective for single-instance representation learning, this is insufficient for heterogeneous dynamical systems because it lacks an explicit mechanism to identify and store shared ODE/DAE patterns across systems.

To improve transferability, LASS-ODE introduces a learnable \emph{common structure hub} (CSH), which acts as a shared repository of reusable dynamic patterns. The CSH token matrix is concatenated with the trajectory embedding matrix \(\tilde{E}\), and both are processed jointly by attention. This allows CSH tokens to interact with system-specific tokens, aggregate shared ODE/DAE structure across systems, and return distilled information to each trajectory representation. Let \(E_{\mathrm{CSH}}\) denote the shared hub tokens. The augmented representation is updated as
\begin{equation}
\begin{aligned}
\label{eq:main_block}
E_0&=(\tilde{E};E_{\mathrm{CSH}}), \\
E_1&=E_0+\mathrm{MHA}\!\big(\mathrm{LN}(E_0),\mathrm{LN}(E_0),\mathrm{LN}(E_0)\big), \\
E_2&=E_1+\mathrm{MoE}\!\big(\mathrm{LN}(E_1)\big),
\end{aligned}
\end{equation}
where \(\mathrm{MHA}\) denotes multi-head attention. In this way, the model captures both system-specific features and reusable cross-system structure. The mixture-of-experts (MoE) block further increases capacity and allows specialization to different dynamic regimes. Stacking \(L\) such blocks yields the final token representation \(\tilde{E}_{\mathrm{final}}\) (bottom right of Fig. \ref{fig:bigpic}).

\noindent\textbf{ODE-constrained decoding.}
Based on \(\tilde{E}_{\mathrm{final}}\), the decoder reconstructs the trajectory through latent continuous-time evolution. Following the latent ODE paradigm, an initial latent state is inferred from the encoded trajectory summary:
\begin{equation}
\boldsymbol{z}_0^{(j)}
\sim
\mathcal{N}\!\big(g_{\mu}(\boldsymbol{h}^{(j)}),\,g_{\sigma}(\boldsymbol{h}^{(j)})\big),
\end{equation}
where $g_{\mu}(\cdot)$ and $g_{\sigma}()$ are learnable neural-network mappings that produce the mean vector and diagonal variance vector, respectively, for the latent initial condition $\boldsymbol{z}_0^{(j)}$.
Instead of using a globally nonlinear vector field, LASS-ODE adopts a token-wise local linear parameterization. For channel \(j\) and token \(k\), the latent dynamics on \(t\in[t_k^{\mathrm{start}},t_k^{\mathrm{end}}]\) are
\begin{equation}
\begin{aligned}
\label{eq:param_estimate}
\dot{\boldsymbol{z}}^{(j)}(t)&=h_{jk}(\boldsymbol{z},t)=A_{jk}\boldsymbol{z}^{(j)}(t)+\boldsymbol{b}_{jk},\\
(A_{jk},\boldsymbol{b}_{jk})&=f_{\mathrm{param}}\!\left(\tilde{E}_{\mathrm{final}}^{(jk)}\right).
\end{aligned}
\end{equation}
Thus, the latent trajectory is generated by a sequence of piecewise linear ODEs, where \(A_{jk}\) and \(\boldsymbol{b}_{jk}\) are predicted from the final token features. The latent trajectory is then obtained by solving
\begin{equation}
\label{eq:solve_linear}
\boldsymbol{z}^{(j)}(t)
=
\mathrm{ODESolve}\!\left(h_{jk}(\boldsymbol{z},t),\boldsymbol{z}_0^{(j)},0,t\right),
\qquad t\in[0,1],
\end{equation}
with \(\boldsymbol{z}^{(j)}(0)=\boldsymbol{z}_0^{(j)}\). This local linear form is especially compatible with GPU-accelerated PyTorch tools, since the latent evolution can be computed efficiently using either general-purpose solvers such as \texttt{odeint} or batched matrix operations such as \texttt{torch.matmul} and \texttt{torch.linalg.matrix\_exp}. Finally, the latent trajectory is mapped back to the physical state space by
\begin{equation}
\hat{\boldsymbol{x}}^{(j)}(t)=W_{\mathrm{dec}}\boldsymbol{z}^{(j)}(t)+b_{\mathrm{dec}},
\end{equation}
where $W_{\mathrm{dec}}$ and $b_{\mathrm{dec}}$ denote the learnable decoder weight matrix and bias vector, respectively. In this way, the decoder preserves a continuous-time interpretation while maintaining scalable training and inference on large heterogeneous trajectory collections.

\vspace{-4mm}
\section{Fine-Tune LASS-ODE to Fit Power System Data}
\label{sec:mix_lora}


Our LASS-ODE model is first \emph{pretrained} on a large set of trajectories spanning diverse ODE/DAE systems across multiple domains. This stage equips the model with general-purpose ODE/DAE simulation capability by learning to estimate token-wise linear ODE parameters in the latent space (see Eq.~\eqref{eq:param_estimate}). Consequently, the latent dynamics in Eq.~\eqref{eq:solve_linear} can be evaluated very efficiently, as they reduce to batched linear ODE integration and matrix operations that are naturally amenable to GPU-accelerated parallel computation. On top of this pretrained foundation model, we further \emph{fine-tune} LASS-ODE using a small set of power-system trajectories so that the model specializes to power-system dynamic behavior and supports accurate power-system time-domain simulation.

\vspace{-5mm}
\subsection{The Specialization-Generalization Tradeoff in Fine-Tuning with Heterogeneous Power Data}
Low-rank Adaptation (LoRA) is a standard parameter-efficient fine-tuning method that adapts a pretrained foundation model by injecting trainable low-rank updates into selected weight matrices, thereby avoiding full-parameter retraining. We focus on LoRA to fine-tune LASS-ODE. In this work, the fine-tuning signals are drawn from domain-specific datasets, namely, power-system ODE/DAE trajectories. However, power-system dynamics form a broad and internally heterogeneous domain: the available trajectories may arise from substantially different subdomains, including EMT signals, synchrophasor-scale responses, and other simulation regimes with distinct temporal resolutions and dynamic characteristics. This heterogeneity induces a specialization-generalization tradeoff during fine-tuning. Incorporating a diverse collection of trajectories can improve domain coverage and support broader generalization, but excessive heterogeneity may obscure subgroup-specific dynamic structure and weaken specialization. Conversely, fine-tuning on a narrowly restricted subset can improve specialization to a particular regime while reducing transfer across the broader power-system domain. To address the specialization-generalization challenge, we propose a mixture-of-LoRA mechanism for adapting LASS-ODE to heterogeneous power-system datasets. The central idea is to automatically identify subgroups in the trajectories and associate them with different low-rank adaptation experts, so that fine-tuning preserves diversity across heterogeneous subgroups while ensuring that each subgroup contributes a coherent adaptation signal.

\vspace{-4mm}

\subsection{LoRA-based Fine Tuning for LASS-ODE }

We consider a pretrained LASS-ODE model with weight matrix
$W \in \mathbb{R}^{d_{\text{out}} \times d_{\text{in}}}$,
for example the query, key, value, or output projection matrices in the
$\mathrm{MHA}$ block in Eq.~\eqref{eq:main_block}, as shown in the top right of Fig. \ref{fig:bigpic}. In the pretraining stage,
the model is trained on large-scale general time-series data, and the resulting
pretrained weight $W$ serves as the foundation for downstream adaptation.

In the fine-tuning stage, LoRA adapts the important weight matrices of
LASS-ODE by adding a low-rank update term that captures domain-specific
information from power-system dynamics. Specifically, LoRA parameterizes the
adapted weight:
\begin{equation}
\begin{aligned}
\tilde{W} &= W + \Delta W,\\
\Delta W &= \frac{\alpha}{r}\B\A,
\label{eq:lora_update}
\end{aligned}
\end{equation}
where $\A \in \mathbb{R}^{r \times d_{\text{in}}}$ and
$\B \in \mathbb{R}^{d_{\text{out}} \times r}$ are trainable low-rank matrices,
$r \ll \min(d_{\text{in}}, d_{\text{out}})$ is the adaptation rank, and
$\alpha$ is a scaling factor. During fine-tuning, the pretrained base weight
$W$ is frozen, and only $\A$ and $\B$ are optimized using power-system dynamic
data. In this way, the model preserves the general knowledge already encoded in
$W$ while injecting domain-specific adaptations through $\Delta W$.

In the testing stage, the learned low-rank update is combined with the frozen
pretrained weight to form the final inference weight
$\tilde{W} = W + \frac{\alpha}{r}\B\A$. The test input is then passed through the
adapted LASS-ODE model using $\tilde{W}$ in place of $W$, without any further
parameter updates. Therefore, LoRA enables the model to leverage both the
general knowledge from pretraining and the power-system-specific knowledge
acquired during fine-tuning, while keeping the number of trainable parameters
small.

\vspace{-4mm}
\subsection{Mixture of LoRA Experts for Heterogeneous Power Data}

To handle heterogeneous dynamic regimes, an intuitive idea is to cluster the trajectories into \(K_{\text{cluster}}\) subgroups and condition the fine-tuning process on this subgroup structure. Here, \(K_{\text{cluster}}\) denotes the number of clusters used to organize the dynamic data according to similarity in their temporal and physical characteristics. As shown in the top right of Fig. \ref{fig:bigpic}, we then introduce \(K_{\text{cluster}}\) LoRA experts, each designed to learn a subgroup-specific fine-tuning update for a corresponding subset of trajectories. In the following, we formalize this process so that the resulting subgroup structure simultaneously maximizes within-group similarity and between-group distinguishability. Based on this structure, the mixture-of-LoRA mechanism is trained to balance specialization and generalization: each expert specializes to a coherent dynamic regime, while the overall model retains transferability across the broader power-system domain. Specifically, we have the following steps.

\textbf{Step 1. K-Means to Cluster Power System Dynamics}. First, we must determine which representation should be used to identify trajectory subgroups. Directly clustering the raw trajectories is generally undesirable, since the raw dynamic data may be noisy, high-dimensional, and heterogeneous in scale and resolution, making subgroup structure difficult to extract reliably. A more suitable choice is to cluster based on a high-level representation that captures the essential dynamic characteristics of each trajectory. Fortunately, in the pretrained LASS-ODE, the encoder in Eq.~\eqref{eq:gru_encoder} produces a summary representation \(\boldsymbol{h}\) of the temporal data across all channels (here we omit the superscript \((j)\)). More importantly, when pretraining is sufficiently extensive, \(\boldsymbol{h}\) is expected to encode the most informative features governing the underlying ODE/DAE process, since the subsequent embedding and prediction modules are all built upon this representation. Therefore, we directly use \(\boldsymbol{h}\) as the high-level feature to represent the input trajectory \(\{\boldsymbol{x}(t_j), t_j\}_{j=1}^N\).

To make the representation scale-invariant for the following grouping, we then normalize the summary feature as
\begin{equation}
\hat{\boldsymbol{h}}=\frac{\boldsymbol{h}}{\|\boldsymbol{h}\|_2}.
\end{equation}

Given the normalized summary features, we partition the trajectories into \(K_{\text{cluster}}\) subgroups by applying \(k\)-means in the normalized feature space. Let \(z_{ik}\in\{0,1\}\) denote the assignment indicator, where \(z_{ik}=1\) means that trajectory \(i\) is assigned to cluster \(k\), and let \(\boldsymbol\mu_k\) denote the corresponding centroid. The clustering problem is formulated as
\begin{equation}
\begin{aligned}
&\min_{\{z_{ik}\},\,\{\boldsymbol\mu_k\}}
\sum_{i=1}^{K_{\text{traj}}}\sum_{k=1}^{K_{\text{cluster}}}
z_{ik}\left\|\hat{\boldsymbol h}_i-\boldsymbol\mu_k\right\|_2^2, \\
& \mathrm{s.t.} z_{ik}\in\{0,1\},\ 
\sum_{k=1}^{K_{\text{cluster}}} z_{ik}=1,
\ i=1,\dots,K_{\text{traj}},
\end{aligned}
\label{eq:kmeans_obj}
\end{equation}
where $K_{\text{traj}}$ is the number of trajectories used in fine-tuning and $z_{ik}$ is the weight. This objective seeks a subgroup partition with high within-group similarity.

Since the features are normalized, minimizing the Euclidean distance in \eqref{eq:kmeans_obj} is equivalent to maximizing the cosine similarity between each trajectory feature and its assigned centroid. In our setting, this angular similarity is especially meaningful because \(\boldsymbol h\) is a learned summary representation produced by the pretrained encoder, and its direction reflects the relative composition of latent dynamic features extracted from the trajectory, whereas its magnitude is more sensitive to overall data scale. Consequently, the angle between two normalized features indicates whether the corresponding trajectories share a similar latent dynamic regime, even when their amplitudes, resolutions, or operating conditions differ. 

A standard solver for Eq. \eqref{eq:kmeans_obj} is Lloyd's algorithm \cite{du2006convergence}. Its modern implementations are highly optimized: for example, scikit-learn reports that Lloyd-style \(k\)-means is fast and memory-efficient in practice, while large-scale vector libraries such as Faiss provide GPU-accelerated clustering routines for dense high-dimensional features.

\textbf{Step 2. Assign Instance to Create The Mixture of LoRA}. After obtaining the cluster centroids, the next step is to associate each trajectory with the corresponding subgroup so that it can guide specialized LoRA fine-tuning. Rather than using a hard assignment, we adopt a soft assignment strategy, which allows each trajectory to contribute to multiple nearby subgroups according to its similarity to their centroids. This design avoids brittle cluster boundaries and enables smoother information sharing across related dynamic regimes. Based on these soft assignment scores, we then construct a trajectory-conditioned mixture of LoRA experts.

Specifically, for a trajectory with normalized summary feature \(\hat{\boldsymbol h}\), we first compute its similarity score to the \(m\)-th centroid \(\boldsymbol\mu_m\) as
\begin{equation}
s_m = \hat{\boldsymbol h}^{\top}\boldsymbol\mu_m,
\qquad m=1,\dots,K_{\text{cluster}}.
\label{eq:routing_score}
\end{equation}
The corresponding soft assignment weight is then defined through a temperature-controlled softmax:
\begin{equation}
\pi_m =
\frac{\exp(s_m/\tau)}
{\sum_{\ell=1}^{K_{\text{cluster}}}\exp(s_\ell/\tau)},
\qquad m=1,\dots,K_{\text{cluster}},
\label{eq:routing_weight}
\end{equation}
where \(\tau>0\) is a temperature parameter controlling the sharpness of the assignment.

We then associate each subgroup with one LoRA expert. Let \((\A_m,\B_m)\) denote the trainable low-rank factors of the \(m\)-th expert. The trajectory-conditioned LoRA update is defined as the weighted combination
\begin{equation}
\Delta W(\boldsymbol h)
= \sum_{m=1}^{K_{\text{cluster}}} \Delta W_m = 
\frac{\alpha}{r}
\sum_{m=1}^{K_{\text{cluster}}}
\pi_m \B_m \A_m,
\label{eq:mix_lora_deltaW}
\end{equation}
where \(r\) is the LoRA rank and \(\alpha\) is the LoRA scaling factor. The final adapted weight for this trajectory is therefore
\begin{equation}
\tilde W(\boldsymbol h)
=
W + \Delta W(\boldsymbol h).
\label{eq:mix_lora_finalW}
\end{equation}
In this way, the pretrained weight \(W\) is adapted through a subgroup-aware mixture of low-rank updates, so that trajectories with similar latent dynamic structure induce similar fine-tuning behavior, while heterogeneous regimes are allowed to activate different LoRA experts.

\textbf{Step 3. Test Computation for New Instance.} In the testing stage, given an unseen trajectory, the pretrained encoder first produces its summary feature \(\boldsymbol h\), from which the soft assignment weights over the \(K_{\text{cluster}}\) experts are computed. These weights determine the corresponding mixture of low-rank updates, which is added to the frozen pretrained weight \(W\) to form the final inference weight \(\tilde W(\boldsymbol h)\). The adapted LASS-ODE model then processes the test input using this trajectory-conditioned weight, thereby enabling input-dependent expert selection and combination at inference time.

Compared with standard LoRA, which applies a single shared low-rank update to all downstream data, the proposed method replaces that uniform adaptation with a trajectory-conditioned mixture of low-rank experts. This design is particularly suitable for heterogeneous power-system datasets, where trajectories may arise from multiple dynamic regimes, temporal resolutions, and simulation environments. By combining subgroup-aware specialization with a shared pretrained backbone, the proposed fine-tuning strategy preserves adaptability across heterogeneous regimes while avoiding the cost of full-parameter retraining.

\vspace{-4mm}

\section{Computational Complexity Analysis}
\label{sec:theorem}

In this subsection, we provide theoretical analyses to demonstrate that (1) the linear ODE formulation in Eqs. \eqref{eq:param_estimate}--\eqref{eq:solve_linear} delivers substantial computational gains, allowing ODE constraints to be enforced in LASS-ODE in a scalable manner suitable for large and foundation-scale training; and (2) the introduction of mixture-of-LoRA in LASS-ODE-Power adds only minimal computational overhead compared with the core inference procedure.

\textbf{Linear Form Brings Significant Computational Gains for Fast Simulations}. Computational efficiency is essential for large-scale power-system dynamic simulation, where thousands of disturbance scenarios or operating conditions may need to be evaluated. Although ODE-based learning models typically introduce additional overhead due to numerical integration, the proposed LASS-ODE formulation reduces the per-step evaluation cost and can achieve lower computational complexity than conventional neural ODE models.
For comparison, we consider a Latent ODE baseline in which the latent derivative is parameterized by a nonlinear multilayer perceptron (MLP), i.e., $\dot{\boldsymbol{z}} = h(\boldsymbol{z};\tilde{E}_{\text{final}}^{(jk)}).$ In contrast, LASS-ODE adopts a piecewise-linear representation $\dot{\boldsymbol{z}} = A_{jk}\boldsymbol{z}+\boldsymbol{b}_{jk}$ in Eq. \eqref{eq:param_estimate}.

For fair comparison, we assume that both the derivative network $h(\cdot)$ and the parameter generator $f_{\text{param}}(\cdot)$ (see Eq.~\eqref{eq:param_estimate}) are $L$-layer MLPs with the same hidden width $K_{\text{width}}$. Let $n_{\text{step}}$ denote the number of numerical integration steps within each token interval.
As summarized in Table~\ref{tab:cost_compare}, LASS-ODE incurs a one-time cost to compute the local linear dynamics $(A_{jk},\boldsymbol{b}_{jk})$, after which each integration step only requires evaluating an affine vector field $A_{jk}\boldsymbol{z}+\boldsymbol{b}_{jk}$. In contrast, the Latent ODE baseline must repeatedly evaluate a nonlinear MLP using the augmented input $[\boldsymbol{z};\tilde{E}_{\text{final}}^{(jk)}]$ at every solver step.

Consequently, the computational complexity of the two approaches differs significantly. LASS-ODE concentrates most computation in a single parameterization stage, while the Latent ODE baseline incurs the full nonlinear evaluation cost at each of the $n_{\text{step}}$ integration steps. As $n_{\text{step}}$ increases, the gap between the two methods grows. Therefore, under comparable solver settings, the proposed formulation can substantially reduce integration overhead and improve scalability for large-scale trajectory simulation.

\begin{table}[ht]
\centering
\caption{ODE integration complexity per token interval. 
$C_{\text{param}}:= d_{\text{model}}K_{\text{width}} + (L-2)K_{\text{width}}^2 + K_{\text{width}}(d_z^2+d_z)$ denotes the one-time cost to compute $(A_{jk},\boldsymbol{b}_{jk})$ from $\tilde{E}_{\text{final}}^{(jk)}$. 
$C_{\text{lin}}:= d_z^2$ is the cost of evaluating the affine vector field $A_{jk}\boldsymbol{z}+\boldsymbol{b}_{jk}$. 
$C_{\text{mlp}}:= (d_z+d_{\text{model}})K_{\text{width}} + (L-2)K_{\text{width}}^2 + K_{\text{width}}d_z$ denotes one evaluation of the nonlinear derivative network $h(\boldsymbol{z};\tilde{E}_{\text{final}}^{(jk)})$.}
\label{tab:cost_compare}
\small
\begin{tabular}{l l l}
\toprule
\textbf{Methods} & \textbf{Derivative} & \textbf{Total cost over $n_{\text{step}}$} \\
\midrule
LASS-ODE & $A_{jk}\boldsymbol{z}+\boldsymbol{b}_{jk}$ 
& $\mathcal{O}\!\left(C_{\text{param}} + n_{\text{step}}\,C_{\text{lin}}\right)$ \\
Latent ODE &$h(\boldsymbol{z};\tilde{E}_{\text{final}}^{(jk)})$ 
& $\mathcal{O}\!\left(n_{\text{step}}\,C_{\text{mlp}}\right)$ \\
\bottomrule
\end{tabular}
\vspace{-4mm}
\end{table}

\textbf{Mixture-of-LoRA Introduces Only Negligible Computational Overhead}. We now analyze how the proposed mixture-of-LoRA fine-tuning affects computational complexity. 
Importantly, the LoRA adaptation operates on the weight matrices of the Transformer backbone (e.g., in the MHA blocks), and does not alter the ODE integration procedure in Eq.~\eqref{eq:solve_linear}.

For a given weight matrix $W \in \mathbb{R}^{d_{\text{out}} \times d_{\text{in}}}$, LoRA introduces a low-rank update $\Delta W = \frac{\alpha}{r}\B\A$, where $\A \in \mathbb{R}^{r \times d_{\text{in}}}$ and $\B \in \mathbb{R}^{d_{\text{out}} \times r}$. 
The additional cost of applying this update to an input vector $\boldsymbol{u} \in \mathbb{R}^{d_{\text{in}}}$ is
\(
\mathcal{O}(r d_{\text{in}} + r d_{\text{out}}),
\)
which is significantly smaller than the original matrix multiplication cost $\mathcal{O}(d_{\text{out}} d_{\text{in}})$ when $r \ll \min(d_{\text{in}}, d_{\text{out}})$.

In the proposed mixture-of-LoRA mechanism, we introduce $K_{\text{cluster}}$ low-rank experts and compute a trajectory-dependent weighted combination of their updates. 
The routing step requires computing similarity scores between the normalized feature $\hat{\boldsymbol{h}}$ and the cluster centroids, which incurs a cost of $\mathcal{O}(K_{\text{cluster}} d_h)$, and $d_h$ is the feature dimension. 
The softmax operation adds negligible overhead.
The resulting low-rank update is a weighted sum of $K_{\text{cluster}}$ expert updates.
A naive implementation would incur $\mathcal{O}(K_{\text{cluster}} r (d_{\text{in}} + d_{\text{out}}))$ cost. 
However, since $K_{\text{cluster}}$ and $r$ are both small constants in practice, this overhead remains modest compared to the base Transformer computation.
Overall, the mixture-of-LoRA mechanism introduces only a linear overhead in $K_{\text{cluster}}$ and does not affect the asymptotic scaling with respect to $n_{\text{step}}$.

\vspace{-4mm}

\section{LASS-ODE-Power for Power Simulations}
\label{sec:power_app}
\vspace{-2mm}

This Section instantiates the measurement-conditioned trajectory simulation framework in Section \ref{sec:problem}, namely interpolation and prediction, for multiple classes of power-system dynamics. As we have introduced the main model LASS-ODE-Power (LOP), we can define $f_{\mathrm{LOP}} \triangleq f_{\mathrm{LASS\text{-}ODE\text{-}Power}}$ and re-emphasize the function of LASS-ODE-Power as follows. Given partial samples of a target signal $\boldsymbol{x}(t)$ up to the last observed time $t_N$, the model returns a continuous-time trajectory over a horizon $[0,t_{\max}]$:
\begin{equation}
\{\hat{\boldsymbol{x}}(t_i)|t_0\leq t_i\leq t_{\max}\} = f_{\mathrm{LOP}}\!\left(\{\boldsymbol{x}(t_i)\}_{i=0}^N\right),
\label{eq:app_io}
\end{equation}
where evaluating $\hat{\boldsymbol{x}}(t)$ for $t \le t_N$ yields interpolation within the observed window, and evaluating $\hat{\boldsymbol{x}}(t)$ for $t>t_N$ yields extrapolation beyond the prefix boundary. 


Next, we specify $\boldsymbol{x}(t)$ in different power-system simulation scenarios. Following the IEEE/CIGRE classification \cite{hatziargyriou2020definition,lindner2025suitable} of power-system dynamic phenomena, we focus on the following representative dynamic processes: generator rotor-angle dynamics, bus-voltage dynamics, system-frequency dynamics, slow converter state dynamics, fast converter state dynamics, and EMT instantaneous voltage and current waveforms at buses, line terminals, and converter terminals.

\vspace{-5mm}

\subsection{Rotor-Angle Transient Dynamics}
\label{sec:app_rotor}
\vspace{-2mm}

Rotor-angle transient stability concerns maintaining synchronism of generators following large disturbances \cite{Kundur2004,alduaij2025beyond}. A canonical representation is the swing-equation-based DAE:
\begin{equation}
\begin{aligned}
\label{eq:rotor_angle}
\dot{\delta}_i &= \omega_i - \omega_s, \\
M_i \dot{\omega}_i &= P_{m,i} - D_i(\omega_i-\omega_s) - P_{e,i}(\delta, V), \\
0 &= g(\delta, V; Y).
\end{aligned}
\end{equation}
Here, $\delta_i$ and $\omega_i$ denote the rotor angle and rotor speed of generator $i$, respectively, and $\omega_s$ is the synchronous speed. 
The parameters $M_i$ and $D_i$ are the inertia and damping coefficients, $P_{m,i}$ is the mechanical input power, and $P_{e,i}(\delta,V)$ is the electrical output power. 
The algebraic constraint $g(\delta,V;Y)=0$ describes the network equations, where $V$ denotes the bus-voltage variables and $Y$ is the network admittance matrix. The given states are $\boldsymbol{x}=[\delta_i,\omega_i]$ and LASS-ODE-Power can help reconstruct the whole trajectory after the disturbance for around $10\sim15$s. In practical repeated stability assessment, transient-stability computations are often performed in batches under periodic refresh cycles. For example, midcontinent independent system operator (MISO) reports online transient-stability analysis on an approximately 15-minute cycle \cite{MISOOnlineDSA2020}, underscoring the operational need for high-throughput trajectory prediction. This makes LASS-ODE-Power a competitive candidate for such applications.

\vspace{-4mm}

\subsection{Bus Voltage Dynamics}
\label{sec:app_voltage}
\vspace{-2mm}

Voltage stability concerns the ability of the system to maintain acceptable bus-voltage magnitudes following disturbances, and is strongly influenced by load-recovery mechanisms and voltage-control dynamics \cite{Kundur2004,Hill1993}. In this work, we instantiate this class under load-change and load-trip events, and denote the monitored bus-voltage channel(s) by $V_i$ for the $i^{th}$ bus. A standard dynamic-load representation for voltage-stability analysis is the exponential recovery load model:
\begin{equation}
\begin{aligned}
\dot{x}_{p,\ell}(t)
&=
-\frac{x_{p,\ell}(t)}{T_{p,\ell}}
+
\bar{P}_{\ell}(t)
\left[
\left(\frac{V_{\ell}(t)}{V_{\ell,0}}\right)^{\alpha_{s,\ell}}
-
\left(\frac{V_{\ell}(t)}{V_{\ell,0}}\right)^{\alpha_{t,\ell}}
\right], \\
\dot{x}_{q,\ell}(t)
&=
-\frac{x_{q,\ell}(t)}{T_{q,\ell}}
+
\bar{Q}_{\ell}(t)
\left[
\left(\frac{V_{\ell}(t)}{V_{\ell,0}}\right)^{\beta_{s,\ell}}
-
\left(\frac{V_{\ell}(t)}{V_{\ell,0}}\right)^{\beta_{t,\ell}}
\right], \\
P_{L,\ell}(t)
&=
\frac{x_{p,\ell}(t)}{T_{p,\ell}}
+
\bar{P}_{\ell}(t)
\left(\frac{V_{\ell}(t)}{V_{\ell,0}}\right)^{\alpha_{t,\ell}}, \\
Q_{L,\ell}(t)
&=
\frac{x_{q,\ell}(t)}{T_{q,\ell}}
+
\bar{Q}_{\ell}(t)
\left(\frac{V_{\ell}(t)}{V_{\ell,0}}\right)^{\beta_{t,\ell}}, \\
0
&=
g\!\big(
V(t),\theta(t);\,
Y,\,
P_G(t)-P_L(t),\,
Q_G(t)-Q_L(t)
\big),
\end{aligned}
\label{eq:voltage_stability_erl}
\end{equation}
where $x_{p,\ell}(t)$ and $x_{q,\ell}(t)$ are the internal recovery states associated with active and reactive power at bus $\ell$; $P_{L,\ell}(t)$ and $Q_{L,\ell}(t)$ are the corresponding active and reactive load demands; $\bar{P}_{\ell}(t)$ and $\bar{Q}_{\ell}(t)$ are event-dependent nominal load levels; $T_{p,\ell}$ and $T_{q,\ell}$ are the active- and reactive-power recovery time constants; $V_{\ell}(t)$ is the voltage magnitude at bus $\ell$ and $V_{\ell,0}$ is its pre-disturbance value; $\alpha_{s,\ell}$ and $\beta_{s,\ell}$ are the steady-state voltage exponents; $\alpha_{t,\ell}$ and $\beta_{t,\ell}$ are the transient voltage exponents; $V(t)$ and $\theta(t)$ collect all bus-voltage magnitudes and phase angles; $Y$ is the network admittance matrix; $P_G(t),Q_G(t)$ denote generator injections; $P_L(t),Q_L(t)$ collect the bus load demands; and $g(\cdot)$ denotes the AC network power-balance equations. Under this formulation, the monitored voltage trajectories $V_l(t)$ are determined implicitly through the evolving load-recovery states and the coupled network algebraic constraints in \eqref{eq:voltage_stability_erl}. 

The defined states are $\boldsymbol{x}=[x_{p,l},x_{q,l}, V_l]$ for LASS-ODE-Power to predict the voltage dynamics. As with rotor-angle studies, operational deployment often involves repeated execution under periodic refresh cycles; for example, MISO reports online voltage-stability calculations on an approximately 6-minute cycle \cite{MISOOnlineDSA2020}.

\vspace{-4mm}
\subsection{Electromagnetic Fault Transients}
\label{sec:app_emt}

Electromagnetic-transient (EMT) simulation resolves instantaneous phase-domain voltages and currents, and is required when waveform-level fault transients or fast converter--control interactions cannot be represented accurately by phasor-domain models \cite{CAISOEMT2020,NESOEMT2025}. We instantiate this class using a Phase-A terminal-voltage waveform during a fault application-and-clearing sequence. A representative EMT-grade network--device model is
\begin{equation}
\begin{aligned}
C_{\sigma(t)} \dot{v}(t) + G_{\sigma(t)} v(t)
&= u_{\sigma(t)}(t) - \AA^\top i_L(t) - \\
&i_{\mathrm{dev}}\!\big(v(t),x_{\mathrm{dev}}(t)\big), \\
L \dot{i}_L(t)
&= \AA v(t) - R i_L(t), \\
\dot{x}_{\mathrm{dev}}(t)
&= f_{\mathrm{dev},\sigma(t)}\!\big(x_{\mathrm{dev}}(t),v(t)\big),
\end{aligned}
\label{eq:emt_mna}
\end{equation}
where \(v(t)\in\mathbb{R}^{n_v}\) is the vector of instantaneous nodal voltages; \(i_L(t)\in\mathbb{R}^{n_L}\) is the vector of currents through the modeled inductive or dynamic branches; \(x_{\mathrm{dev}}(t)\in\mathbb{R}^{n_x}\) collects the fast internal states of converters, controllers, machines, or other dynamic devices; \(C_{\sigma(t)},G_{\sigma(t)}\in\mathbb{R}^{n_v\times n_v}\) are the topology-dependent capacitance and conductance matrices; \(L,R\in\mathbb{R}^{n_L\times n_L}\) are the branch inductance and resistance matrices; \(u_{\sigma(t)}(t)\in\mathbb{R}^{n_v}\) is the nodal source-injection vector; and \(i_{\mathrm{dev}}(t)\in\mathbb{R}^{n_v}\) is the nodal current injection from nonlinear or controlled devices. The matrix \(\AA\in\mathbb{R}^{n_L\times n_v}\) is the reduced branch-to-node incidence matrix associated with the chosen branch orientations, so that \(\AA v(t)\) is the vector of branch voltage drops and \(\AA^\top i_L(t)\) is the corresponding nodal current injection. The switching mode \(\sigma(t)\in\mathcal{S}\) is piecewise constant and encodes fault application, fault clearing, breaker operations, and other topology changes. The function \(f_{\mathrm{dev},\sigma(t)}(\cdot)\) describes the device-state evolution under mode \(\sigma(t)\). Finally, we can define the system state $\boldsymbol{x}=[v,i_L]$ as input to LASS-ODE-Power.

\begin{figure*}[t]
\centering
\includegraphics[width=\linewidth]{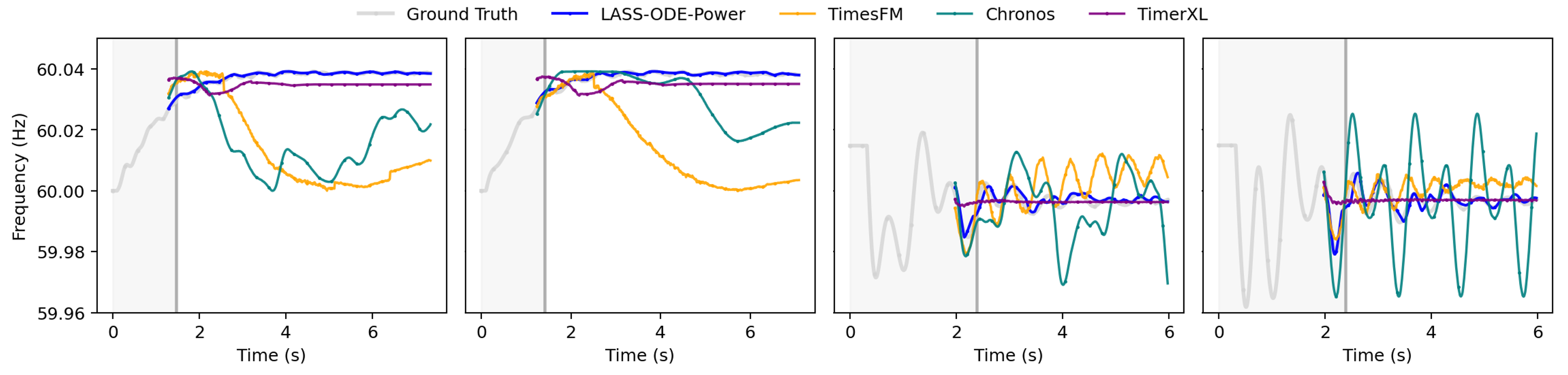}
\caption{Frequency trajectory prediction in the IEEE 39-bus system. The left two panels show over-frequency cases and the right two show under-frequency cases, under \(\pm 0.1\)~p.u. load changes. The left two use 20\% of each trajectory as input, and the right two use 40\%. Gray denotes ground truth, colored curves denote predictions, and the vertical line separates the observed and forecast parts.}
\label{fig:benchmarkcomparisonfrequency}
\vspace{-4mm}
\end{figure*}

\vspace{-4mm}


\begin{figure*}[t]
\centering
\includegraphics[width=\linewidth]{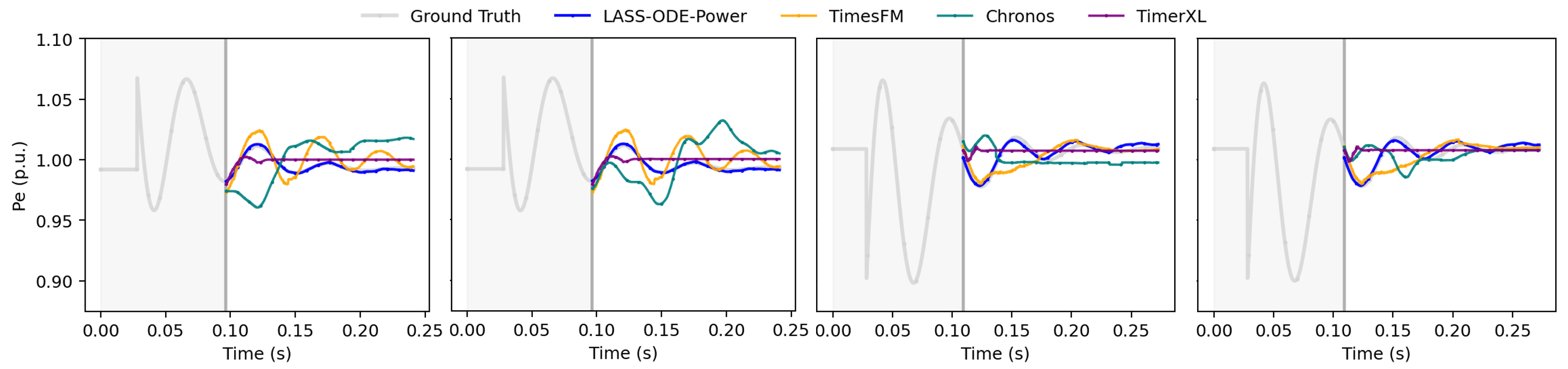}
\caption{Active-power trajectory prediction under post-event slow converter instability. All four cases use 40\% of each trajectory as the observed input. The two on the left show active-power increase cases, and the two on the right show active-power decrease cases.}
\label{fig:benchmarkcomparisonfastconverter}
\vspace{-6mm}
\end{figure*}

\section{Numerical Validation}
\label{sec:results}
\vspace{-2mm}

This Section evaluates the proposed LASS-ODE-Power framework on multiple power-system dynamic simulation tasks. 

\vspace{-5mm}
\subsection{Experimental Settings}
\vspace{-2mm}
We select the following cases for experiments: \textbf{Frequency dynamics}. We study frequency dynamics in the multiple standard power systems (e.g., IEEE 39-bus system) by applying load disturbances ($\pm \{0.1, 0.2, 0.3, 0.4\}$~p.u.\ ) to create over-frequency and under-frequency disturbances. We record generator-frequency trajectories at generator buses. \textbf{Converter power dynamics}. Similarly, we conduct simulations at multi-converter grid-connected systems in \cite{huang2022impacts}, such as a modified IEEE 39-bus system. We also examine post-event converter stability by recording active-power trajectories at buses with converters following similar load disturbances. We then extract active-power trajectories for tests. 
\textbf{Generator power dynamics:} We use multi-machine electromechanical power-system models and simulate short-circuit disturbances followed by fault clearing (e.g., three-phase faults with varying clearing times) and record generator electrical power trajectories. \textbf{Fast waveform dynamics:} We use three-phase EMT models and apply faults such as three-phase and single-line-to-ground faults. Voltage waveforms are recorded. To provide contexts information to indicate event clearing, we also give a trajectory of event clearing time (a constant curve) as an additional input dimension to all methods. 



We compare LASS-ODE with TimesFM \cite{das2023decoder}, Chronos \cite{ansari2024chronos}, and TimerXL \cite{liu2024timer}. These models are strong time-series baselines. For all methods, we collect 1 GB of simulated data for fine-tuning. We use an isolated test data set with similar event types but different conditions (such as load disturbances) and physical parameters in the DAE to test. The mean square error (MSE) is reported and compared.

\begin{table*}[t]
\centering
\caption{MSE (\(\times 10^{-2}\)) for IEEE 39-bus frequency prediction under a single \(0.1\)~p.u. load decrease (over-frequency) and a single \(0.1\)~p.u. load increase (under-frequency). Bus 30 is a fine-tuned case, where some of its training trajectories are included in the fine-tuning set, while Buses 31--33 are zero-shot cases, whose trajectories are never seen during training or fine-tuning.}
\small
\setlength{\tabcolsep}{5pt}
\renewcommand{\arraystretch}{1.15}
\begin{tabular}{llcccccccc}
\toprule
\multirow{2}{*}{Bus} & \multirow{2}{*}{Setting} & \multicolumn{4}{c}{Over-frequency MSE ($\times 10^{-2}$)} & \multicolumn{4}{c}{Under-frequency MSE ($\times 10^{-2}$)} \\
\cmidrule(lr){3-6}\cmidrule(lr){7-10}
 &  & LASS-ODE & TimesFM & Chronos & TimerXL & LASS-ODE & TimesFM & Chronos & TimerXL \\
\midrule
Bus 30 & Fine-tuning & \textbf{0.09} & 173.60 & 50.05 & 21.60 & \textbf{0.70} & 33.56 & 85.31 & 3.69 \\
Bus 31 & Zero-shot & \textbf{0.12} & 166.40 & 53.00 & 26.29 & \textbf{3.37} & 29.88 & 43.64 & 5.00 \\
Bus 32 & Zero-shot & \textbf{0.14} & 177.71 & 41.54 & 24.38 & \textbf{2.55} & 23.90 & 93.65 & 4.67 \\
Bus 33 & Zero-shot & \textbf{0.37} & 171.06 & 56.40 & 8.40 & \textbf{3.48} & 40.91 & 55.68 & 4.80 \\
\bottomrule
\end{tabular}

\label{tab:benchmark-mse}
\vspace{-3mm}
\end{table*}


\begin{figure}[t]
\centering
\includegraphics[width=\linewidth]{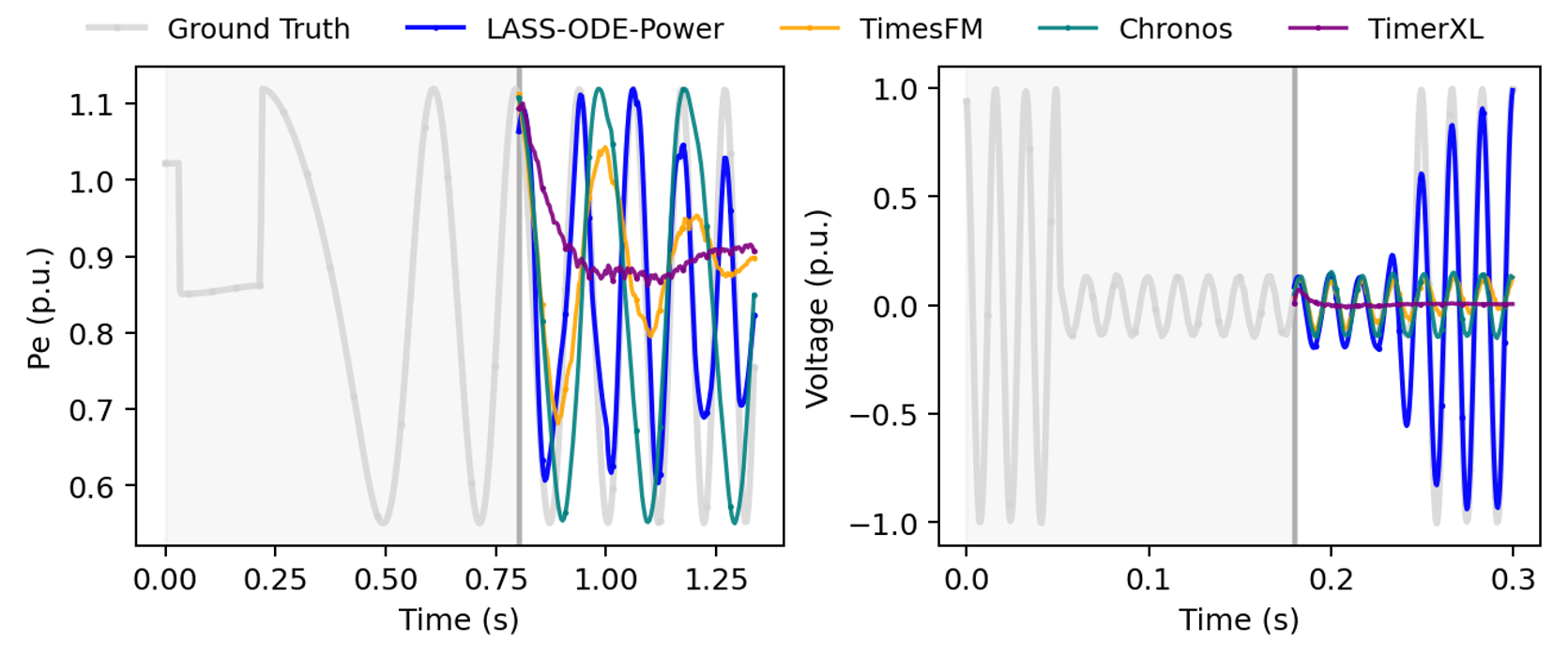}
\caption{Fault-driven trajectories with 60\% observed input. The left one shows generator electrical power during a transient-stability event, and the right one shows the Phase-A voltage waveform during a three-phase fault.}
\label{fig:benchmarkcomparisonthreephase&generator}
\vspace{-6mm}
\end{figure}

\vspace{-4mm}

\subsection{Trajectory Prediction Visualization}
\label{sec:sim_visual}
\vspace{-2mm}

Next, we visualize representative predictions in Figs. \ref{fig:benchmarkcomparisonfrequency} to \ref{fig:benchmarkcomparisonthreephase&generator}. These foundation models are applied to trajectory prediction across very different power-system dynamics. Under different input ratios, the blue curve from LASS-ODE-Power most closely matches the gray ground-truth curve. This is because LASS-ODE-Power captures the underlying DAE evolution and constrains prediction through DAE parameterization. In addition, mixture-of-LoRA enables rapid adaptation to system-specific DAE parameters for accurate prediction. By contrast, other foundation models mainly capture generic time-series patterns, but do not model the underlying DAEs well.  

\vspace{-4mm}

\subsection{Zero-shot and Fine-tuned Results}
\label{subsec:lora_fine_tuning}
\vspace{-2mm}

Table \ref{tab:benchmark-mse} reports the zero-shot and fine-tuned results of LASS-ODE and the baselines on different power-system simulation tasks. Across both disturbance types and all observation ratios, LASS-ODE achieves the lowest prediction error. Similar gains are also observed on the other datasets.

\vspace{-4mm}
\subsection{Inference Time}
\vspace{-2mm}

\begin{table}[h]
\centering
\caption{Test time (s) comparison.}
\begin{tabular}{l c}
\hline
Model & Time Length \\
\hline
LASS-ODE & 0.379 \\
TimesFM  & 0.581 \\
Timer-XL & 0.315 \\
Chronos  & 1.707 \\
\hline
\end{tabular}
\label{tab:time_length}
\end{table}

Table \ref{tab:time_length} reports the test time of $64$ trajectories. Thanks to the sparse mixture-of-experts backbone and the piecewise-linear ODE decoder, LASS-ODE-Power achieves moderate latency and can perform trajectory prediction in near real time. This efficiency makes the framework suitable for repeated prediction tasks in operational studies.

\vspace{-4mm}
\section{Conclusion}
\label{sec:conclusion}
\vspace{-2mm}

This paper introduces LASS-ODE-Power, a foundation model for general-purpose power-system time-domain simulation. Given a prefix of online measurements, LASS-ODE-Power predicts the future system evolution. It is a large-scale pretrained model that can be adapted, using only data, to different simulation tasks, such as frequency stability, voltage stability, and EMT simulation, without requiring knowledge of the underlying ODE/DAE parameters. Specifically, we show how to obtain LASS-ODE-Power by fine-tuning our prior model, LASS-ODE, on power-system datasets. To handle the multi-timescale and heterogeneous nature of these datasets, we propose a novel mixture-of-LoRA strategy that performs adaptation across different groups of power-system dynamics. Finally, extensive experiments on diverse power-system tasks demonstrate that LASS-ODE-Power consistently achieves the lowest prediction error and fast inference.

\vspace{-4mm}

\bibliographystyle{IEEEtran}
\bibliography{reference}

\end{document}